\documentclass[pre,preprint,showpacs]{revtex4-1}

\usepackage{graphicx}
\usepackage[latin1]{inputenc}
\usepackage{amsfonts}
\usepackage{amsmath,amssymb}
\usepackage{bm}
\usepackage{cancel}
\usepackage{cleveref}
\usepackage{subfigure} 
\usepackage{bbold}
\usepackage{caption}
\usepackage{graphicx,color}
\usepackage{pgf,tikz}
\usetikzlibrary{shapes,arrows}
\usepackage[export]{adjustbox}

\begin{document}

\title{Hierarchical topological clustering}

\author{
  Ana Carpio$^{1}$, Gema Duro$^{2}$
}

\affiliation{$^1$Facultad de Ciencias Matem\'aticas, Universidad Complutense 
 de Madrid, Plaza de Ciencias 3, Madrid 28040, Spain}
\affiliation{$^2$Facultad de Ciencias Econ\'omicas y Empresariales, Universidad 
   Aut\'onoma de Madrid, Tomas y Valiente 5, Madrid 28049, Spain }
\date{\today}

\begin{abstract}
Topological methods have the potential of exploring data clouds without 
making assumptions on their the structure. Here we propose a hierarchical 
topological  clustering algorithm that can be implemented with any 
distance choice.  The persistence of outliers and clusters of arbitrary
shape is inferred from the resulting hierarchy. We demonstrate the potential 
of the algorithm on selected datasets in which outliers play relevant roles, 
consisting of images, medical and economic data. These methods
can provide meaningful clusters in situations in which other techniques
fail to do so.
\end{abstract}

%\begin{keywords}
%clustering methods, topological data analysis, outliers, hierarchies
%\end{keywords}

\maketitle

\section{Introduction}
\label{sec:intro}

Cluster analysis intends to categorize objects in groups according to
their affinity. Applications include pattern recognition, image processing, 
market research and data analysis.  
All clustering procedures are affected by the presence of noise
and outliers. Noise is random error. An outlier is any point in the 
dataset  that deviates noticeably when compared with other points. 
While extreme values caused by errors are just noise and can
be discarded, meaningful outliers are part of the data. They may
indicate the presence of different mechanisms or exceptional 
elements. Here, we propose a hierarchical topological clustering (HTC)
algorithm which is able to identify clusters of arbitrary geometry
and distinguish relevant outliers in an automatic way.
Our algorithm is based on persistent homology, a tool used in topological
data analysis to identify data features that persist across scales. 

Clustering techniques typically group elements in such a way that the 
items in the same group (a cluster) are closer to each other \cite{clusters,
survey} than to the items belonging to other groups (in some sense). 
An increasing variety of clustering strategies which define clusters in 
different ways  are being proposed for different purposes, see
\cite{survey,ams} for a review.
Popular approaches are based on partitions, hierarchies, densities
or distributions.  Clustering algorithms based on partitions  group
data in a prefixed number of sets seeking to minimize the distance
to cluster centers. K-medoids \cite{clusters} and K-means \cite{kmeans} 
fall in this cathegory. These methods fail to identify non convex clusters 
because observations are always assigned to the closest center.
% kmedoids any measure of dissimilary, pairwise, center
% must be a data point
% kmeans uses always euclidean distance
Hierarchical clustering \cite{clusters} builds a hierarchy of clusters
relying on a linkage criterion that specifies the dissimilarity between 
sets in terms of the pairwise distance between observations
in the set.  Its agglomerative variant considers each object a
single cluster initially.  Next, pair of clusters merge until a final large
cluster containing all is formed. This procedure provides a tree-like
representation of the data, named dendrogram.
% works with any distance
While K-means tries to optimize a global objective (the variance of the 
clusters) and reaches a local optimum, agglomerative hierarchical 
clustering searches the best step at each cluster fusion, resulting in 
a potentially suboptimal solution subject to many arbitrary choices.
Nevertheless, K-means with random initialization can lead to different 
results when run several times on  the same dataset.
Distribution-based algorithms aim to emulate the observed data 
structure as a superposition of pre-established probability
distribution functions \cite{em}. Success depends on the ability
of the proposed probability distribution to reproduce the data.
The local density of data points is directly exploited by density-based 
spatial clustering of applications with noise (DBSCAN) \cite{dbscan}.
This method assigns clusters to point clouds of arbitrary shape whose 
density surpasses a selected threshold. The remaining points are
considered outliers or noise. However, choosing meaningful thresholds 
is often a nontrivial task. More recent clustering algorithms combine 
density and distance. An approach  based on the idea that that cluster 
centers are distinguished by a higher density and by a  certain  distance
from other points with high densities is explored in  \cite{science}. 
Other novel clustering strategies rely on graph theory, grid or network 
structures, fractal theory, statistical or neural network learning, and kernels 
\cite{survey,ams}.

We focus here on clustering algorithms based on the idea of shape.
In the last years, topological methods, which refers to methods of 
exploiting topological features in data (such as connected components, 
voids, tunnels, etc.) have attracted considerable interest \cite{tclassifying,
tnetworks,tgeometrical}. Different ways to incorporate topological features
give rise to different topological clustering algorithms. 
A common trend  nowadays is to characterize point clouds by their 
persistence diagrams \cite{tda2}, calculate Wasserstein distances 
\cite{twasserstein2} between them and apply standard clustering algorithms 
to group datasets by their  similarity. Applications include network clustering 
\cite{tnetworks}, measurement of function dissimilarity \cite{tgeometrical},  
time series clustering \cite{tseries}, clustering of interfaces, aggregates 
and  patterns  \cite{plos, taggregates, tpatterns}, as well as studies of the 
structure of swarms and flocks \cite{tswarm, swarms}, to quote a few.
Our aim here is not to group datasets or points clouds according to
their persistence diagrams, but to cluster the elements of the datasets
themselves according to basic topological features. Thus, we use
homology information in a different way. 
A related idea is discussed in \cite{ams} together with a different
computational algorithm formulated on networks whose points have
assigned densities. Here, we introduce a different algorithm that focuses 
just of point clouds and distances to calculate the $H_0$ homology 
\cite{tda3} while keeping track of the individual elements of each 
component during the process of constructing a hierarchy.
Like DBSCAN, the algorithm we propose here is able to detect clusters 
of arbitrary shape and extreme values. However, it does not require blind
choices of parameters. The persistence of outliers and clusters can 
be inferred from the resulting hierarchy as we will demonstrate next on 
selected datasets. 
As usual in persistent homology studies, once a distance that measures 
the similarity of the items is chosen, we construct a nested one-parameter  
of simplicial complexes. Then, 
%instead of comparing the simplicial complexes 
%of different point clouds as described by their persistence diagrams, 
we use the simplicial complex to define a clustering hierarchy. For each
value of the governing parameter, the disjoint components of the simplicial
complex corresponding to that parameter define clusters of arbitrary shape. 
They merge into larger clusters as we vary the governing parameter and 
evolve to a different simplex of the family. Clusters and outliers arise 
naturally in this way. Dominant  clusters usually mark dense areas, while 
outliers may represent meaningful rare elements. This method is not too 
sensitive to noise.
%the choice of metric and to noise. 

The paper is organized as follows. First, we explain the algorithm 
and illustrate the results in a geometrical example, a fragmented
interface representing the invasion of healthy tissue by malignant
cells. While other standard algorithms (Kmeans, hierarchical 
clustering with the highest cophenetic correlation coefficient, 
DBSCAN) fail to identify clusters with a clear meaning in the 
application considered, our topological hierarchical clustering 
algorithm is able to distinguish the main interface between healthy
and cancerous tissue as well as detached islands of malignant cells
that have migrated into the healthy cells.
%The hierarchy of clusters we obtain differs from the standard ones.
We are particularly interested in the potential of this algorithm to
identify meaningful outliers, which we test on datasets from image 
analysis, gene data research and economy studies. 
The Appendix collects technical details on the clustering
algorithms used for comparison, non Euclidean distantes used
for image and gene data, as well the compression procedure employed
in the images tests.

\section{Materials and methods}
\label{sec:methods}

Our approach is based on ideas borrowed from topological data analysis, 
more precisely, from persistent homology.
Persistent homology is an adaptation of homology to clouds of point data 
\cite{tda2,tda1} which provides a method for computing their topological 
features at different spatial resolutions, using any chosen distance.
Highly persistent features span over a wide range of spatial scales, being 
more likely to represent true features of the data  under study than to 
constitute artifacts of sampling, noise, or parameter choice \cite{tda3}. 

\paragraph{Hierarchical topological clustering.}
%\label{sec:htc}
To find the persistent homology of a cloud of point data $X$, we must 
first represent it in terms of simplicial complexes.  Once a metric $d$ to 
measure the distance between data points has been selected, we generate 
a filtration of the simplicial complex. A filtration is a nested sequence of 
increasingly bigger subsets. A typical choice is the Vietoris-Rips filtration 
\cite{tda1,tda2}.  Given a finite point cloud $X$, {\it the Vietoris-Rips filtration} 
${\rm VR}(X,r)$ is a simplicial complex is defined for $r >0$ and constructed 
as follows
\begin{itemize}
\item The set of data points $X$ are the vertices.
\item Given vertices $x_1$ and $x_2$, the edge $[x_1,x_2]$
is included in ${\rm VR}(X,r)$ if the distance $d(x_1,x_2) \leq r$. 
\item If all the vertices of a higher dimensional simplex are included
in ${\rm VR}(X,r)$, the simplex belongs to ${\rm VR}(X,r)$.
\end{itemize}
According to these conditions ${\rm VR}(X,r) \subset {\rm VR}(X,r')$ 
whenever $r \leq r'$.
%We can work with the Euclidean distance, or any of the distances defined 
%in Section \ref{sec:distances}. 
While constructing a Vietoris-Rips filtration, the data points are connected 
by edges as $r$ grows, forming connected components. For each 
value of $r$, the number of components is the Betti number $b_0(r)$. 
If we identify components with clusters, this procedure induces a 
{\it hierarchical topological clustering strategy} (HTC):
\begin{itemize}
\item We choose the distance $d$ and the range for the filtration parameter 
$r$ 
\begin{eqnarray*}\
\left\{0, {r_{\rm max} \over M}, {2r_{\rm max} \over M},
\ldots, {(M-1) r_{\rm max} \over M}, r_{\rm max} \right\},
\end{eqnarray*}
where $r_{\rm max} = {\rm max}_{x,y \in X}d(x,y)$ (the diameter of the 
point cloud) and ${r_{\rm max} \over M} \sim {\rm min}_{x,y \in X, 
x \neq y}d(x,y)$ (the minimum distance between points), for instance.
\item For $r =0$, each data point constitutes a component/cluster 
itself. 
\item For each value $r$ we construct edges joining points at distance
smaller that $r$. Each set of points connected by edges defines
a cluster. 
\item  When $r$ reaches the diameter of the point cloud, all possible 
edges form and we are left with a single component/cluster. 
\end{itemize}
Persistently isolated points or clusters are last to join and provide 
natural candidates to being outliers (single or collective).
Selecting at a definite filtration value $r$, the Betti number 
$b_0(r)$ provides the number of clusters at that level. The whole 
family of clusters as $r$ varies provides the topological hierarchy of
clusters.
Higher order levels of the Vietoris-Rips filtration are not needed for
this clustering procedure, the $H_0$ level (homology) is enough.

\paragraph{HTC Algorithm.}
%\label{sec:algorithm}
%A number of open access platforms are available for homology
%studies of point clouds, such as Perseus or Javaplex. 
Since we only use the $H_0$ homology, the algorithm can be implemented
in a straightforward way. We propose the following process:
\begin{itemize}
 \item Given a point cloud $X= \{ x_1,\ldots, x_N \}$  formed by $N$ 
points, calculate the matrix of distances for the selected metric $d$.
 \item Calculate $r_{\rm max} = {\rm max}_{x,y \in X}d(x,y)$ and
 $r_{\rm min} = {\rm min}_{x,y \in X, x \neq y}d(x,y)$.
 \item Choose $M = [{r_{\rm max} \over r_{\rm min}}]$ and set the step 
 $h = {r_{\rm max}\over M} > r_{\rm min}$.
 \item Construct the grid of increasing filtration values $r_m=m \, h$, 
 $m=0,1,\ldots, M$.
 \item For $r_0=0$, define $N_0=N$ clusters formed by a single 
point each.
\item For $m=1,\ldots M$
 \begin{itemize}
   \item Start with clusters $C_i$, $i=1, \ldots, N_{m-1}$ and set
   $N_m=N_{m-1}$. 
 %\item calculate the matrix of point links $L_k$, $L_k(i,j)=1$ if $d(x_i,x_j) < r_k$,
%$L_k(i,j)=0$, otherwise,
   \item Calculate the $N_m \times N_m$ matrix of cluster links 
    $L_m$, $L_m(i,j)=1$ if $d(x_i,x_j) 
    < r_m$, for some $x_i \in C_i$ and some $x_j \in C_j$, $L_m(i,j)=0$, 
    otherwise,
  \item If $m=1$ or $m>1$ and $L_m \neq L_{m-1}$  \\
    %\begin{itemize}
    * Consider the first cluster $C$, \\
     \hskip 2mm .  mark $C$ as visited, \\
     \hskip 2mm . list the clusters $C'$ linked to it ($L_m(C,C')=1$), \\
     \hskip 2mm .  for each unvisited cluster $C'$, mark as visited,  
        merge to $C$, list the clusters $C''$ linked to the union ($L_m(C,C'')=1$), 
        and repeat recursively until all linked unvisited clusters merge with $C$. \\
   * For each unvisited cluster $C$, repeat the procedure, until all
    clusters are visited. \\
   * Set $N_m$ equal to the final number of clusters. \\
   * If $N_m=1$ stop.
   % \end{itemize}
  % otherwise nothing is done
  \end{itemize}
  \item Output: the hierarchy of clusters and their elements for each filtration value.
\end{itemize}
A simple way to calculate the matrix of cluster links $L_m$ is to calculate
previously the $N_m \times N_m$ matrix of point links $P_m$ given by $P_m(i,j)=1$ 
if $d(x_i,x_j) < r_m$, and $P_m(i,j)=0$, otherwise. Then, 
$L_m(i,j) = 0$ if $P_m(i',j')=0$ for all $x_{i'}\in C_i$ and $x_{j'}\in C_j$, otherwise
$L_m(i,j) = 1$. 
This algorithm allows us to list the individual elements of each cluster for each 
filtration value and keep track of the clusters it belongs to. Figure \ref{fig1} visualizes
the process. 

The complexity of this algorithm is at most $MN^2$, with lower bound of order
$\Pi_{m=1}^M N_m^2$, $N_m$ being the number of clusters at filtration value $r_m$.
Notice that no operations are performed as the filtration value grows unless
the matrix of cluster links changes. No operations are performed either once a single 
cluster is formed. This algorithm is efficient for small and moderate datasets. As the 
size of the dataset grows, density based algorithms could become an option
\cite{ams}.

%A Matlab routine implementing this algorithm is included as 
%Supplemental Material. The input data are the filtration grid and the matrix of distances
%between point. The output is a matrix containing a cluster per row, and a vector
%indicating the length of each cluster.

%O(n^2) in the worst case for each filtration value we fix
%HC O(kn^2) k= numero clusters
%Check neighbor of each point
%Minpts = 1  check compare with dbscan Mtps=1 for all
%the grid of r
    
%

\tikzstyle{block} = [rectangle, draw, fill=green!10, node distance=1.5cm,
    text width=12em, text centered, rounded corners, minimum height=2em]
\tikzstyle{circle} = [ellipse, draw, fill=blue!10, node distance=4.0cm,
    text width=8em, text centered, minimum height=4em]    
\tikzstyle{triangle} = [diamond, shape aspect= 2, draw, fill=red!10, text width=6em, text centered, 
    minimum height=2em]      
\tikzstyle{key} = [diamond, shape aspect= 2, draw, fill=red!10, text width=4em, text centered, 
    minimum height=2em]  
\tikzstyle{line} = [draw, -latex']

\begin{figure}[!hbt]
\centering \scriptsize
\begin{tikzpicture}[auto] %node distance = 2cm, 
    % Place nodes
    \node [circle] (data) 
             {Point cloud \\
              $x_n$, $n=1,\ldots,N$
              Distance matrix
               };
    \node [circle, right of=data] (filtration) 
             {Filtration grid $r_m$\\
              $m=0,\ldots,M$};
     \node [circle, right of=filtration] (points)
             {Initial clusters  \\
              $N_0=N$, $C_i=x_i$, \\
              $i=1,\ldots,N_0$ \\
               };
     \node [%shape aspect= 2, 
               key,   node distance = 2.2cm, below of=points] 
              (iteration)  {for $m=1,\ldots M$
               };
     \node [block, node distance = 4.0cm, left of=iteration] (clusters) 
              {$C_i$,  $i=1,\ldots,N_{m}=N_{m-1}$  
              };    
     \node [block, node distance = 4.2cm, left of=clusters] (links) 
             {$N_m \times N_m$ matrix of cluster \\ links $L_m$  
             };                 
     \node [%shape aspect= 2, 
                triangle, node distance = 1.6cm, below of=links] (recursion) 
             {$L_m \neq L_{m-1}?$
              };       
     \node [block, text width=15em, node distance = 2.0cm, below right of=recursion] (unvisited) 
             {List unvisited clusters $U$
              };   
     \draw (recursion) -- (unvisited) node [midway, above, sloped] (TextNode) {Yes}; 
     \draw (recursion) -- (iteration) node [midway, below, sloped] (TextNode) {No}; 
     \node [block, text width=15em, node distance =1.0cm, below of=unvisited] (delete) 
             {Select first $C$, detete from $U$
              };        
     \node [block, text width=15em, node distance = 1.4cm, below of=delete] (linked) 
             {List unvisited clusters $C'$ linked to $C$ \\
              $L_m(C,C')=1$, $C' \in U$
              };    
     \node [block, text width=15em, node distance = 1.6cm, below of=linked] (merge) 
             {Select first $C'$, detete from $U$, \\
              Merge $C'$ with $C$ to get $C$
              };     
     \node [block, text width=15em, node distance = 1.6cm, below of=merge] (unvisited') 
             {List unvisited clusters $C''$ linked to new $C$ \\
              $L_m(C,C'')=1$, $C'' \in U$
              };  
     \node [block, text width=15em, node distance = 3.2cm, below of=merge] (recursion') 
             {Repeat recursively until all clusters are visited
              }; 
     \node [circle, node distance = 1.9cm, below of=recursion'] 
            (clusters') 
             {final $N_m$ clusters, \\
             list of elements of each cluster
              }; 
    \node [%ñshape aspect= 2, 
               triangle,  node distance = 4.0cm, right of=clusters'] (continue) 
             {$N_m>1$? 
              }; 
   \node [%shape aspect= 2, 
               key, node distance =  3.0cm, right of=continue](end)  
             {Stop
              }; 
     \draw (continue) -- (iteration) node [midway, above, sloped] (TextNode) {Yes}; 
     \draw (continue) -- (end) node [midway, below, sloped] (TextNode) {No};

    % Draw edges
    \draw [line] (data) -- (filtration);
    \path [line] (filtration) -- (points);
    \path [line] (points) -- (iteration);
    \path [line] (iteration) -- (clusters);
    \path [line] (clusters) -- (links);
    \path [line] (links) -- (recursion);
    \path [line] (recursion) -- (unvisited);
    \path [line] (recursion) --(iteration);
    \path [line] (unvisited) -- (delete);
    \path [line] (delete) -- (linked);
    \path [line] (linked) -- (merge);
    \path [line] (merge) -- (unvisited');
    \path [line] (unvisited') -- (recursion');
    \path [line] (recursion') -- (clusters');
    \path [line] (clusters') -- (continue);
    \path [line] (continue) -- (iteration);
    \path [line] (continue) -- (end);
\end{tikzpicture}
\caption{{\bf Scheme of the HTC algorithm.}}
\label{fig1}
\end{figure}
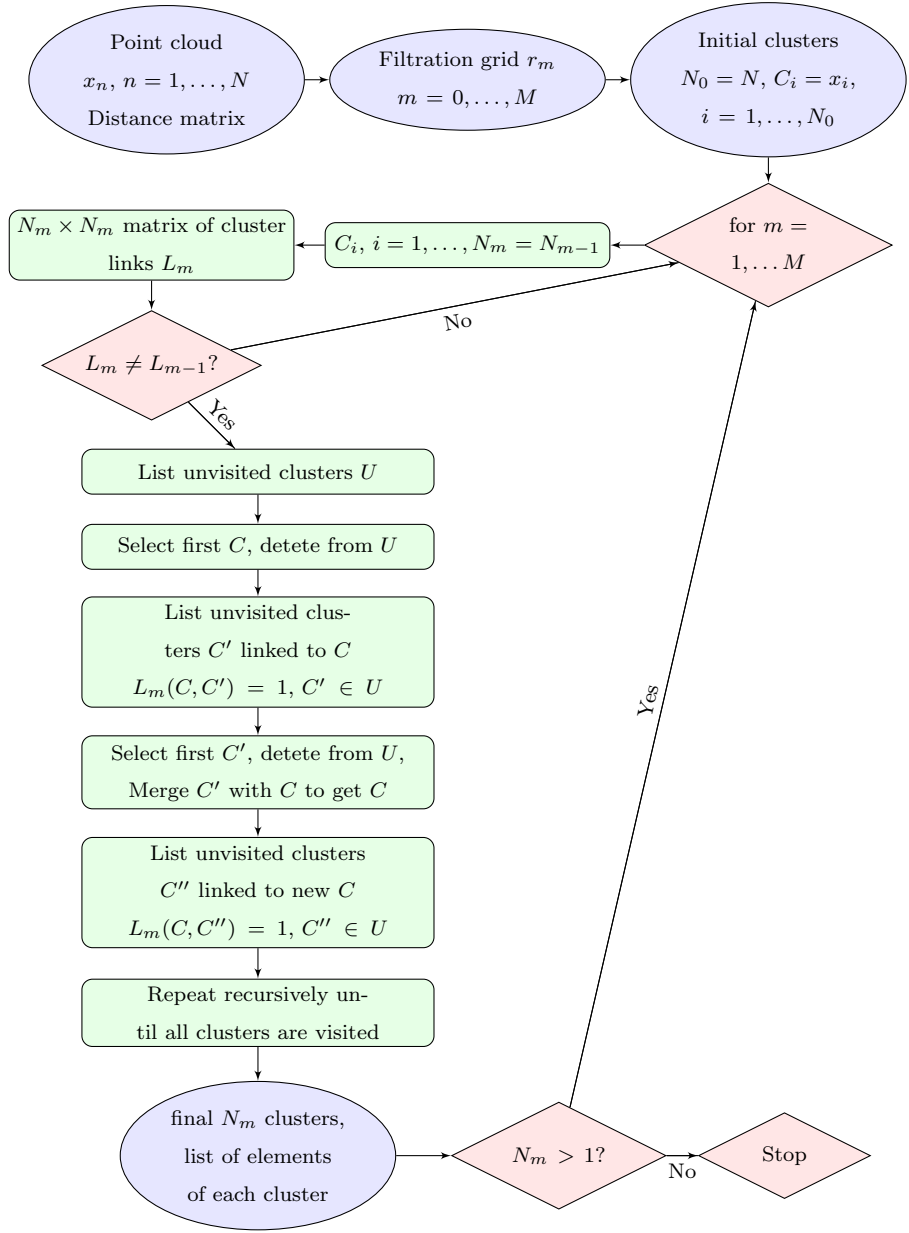

\section{Results and discussion}
\label{sec:results}

We illustrate the performance of the algorithm on several datasets, taken for 
geometry studies, image analysis, trade
and gene datasets next.

\begin{figure}[!hbt] \centering
\includegraphics[width=16cm]{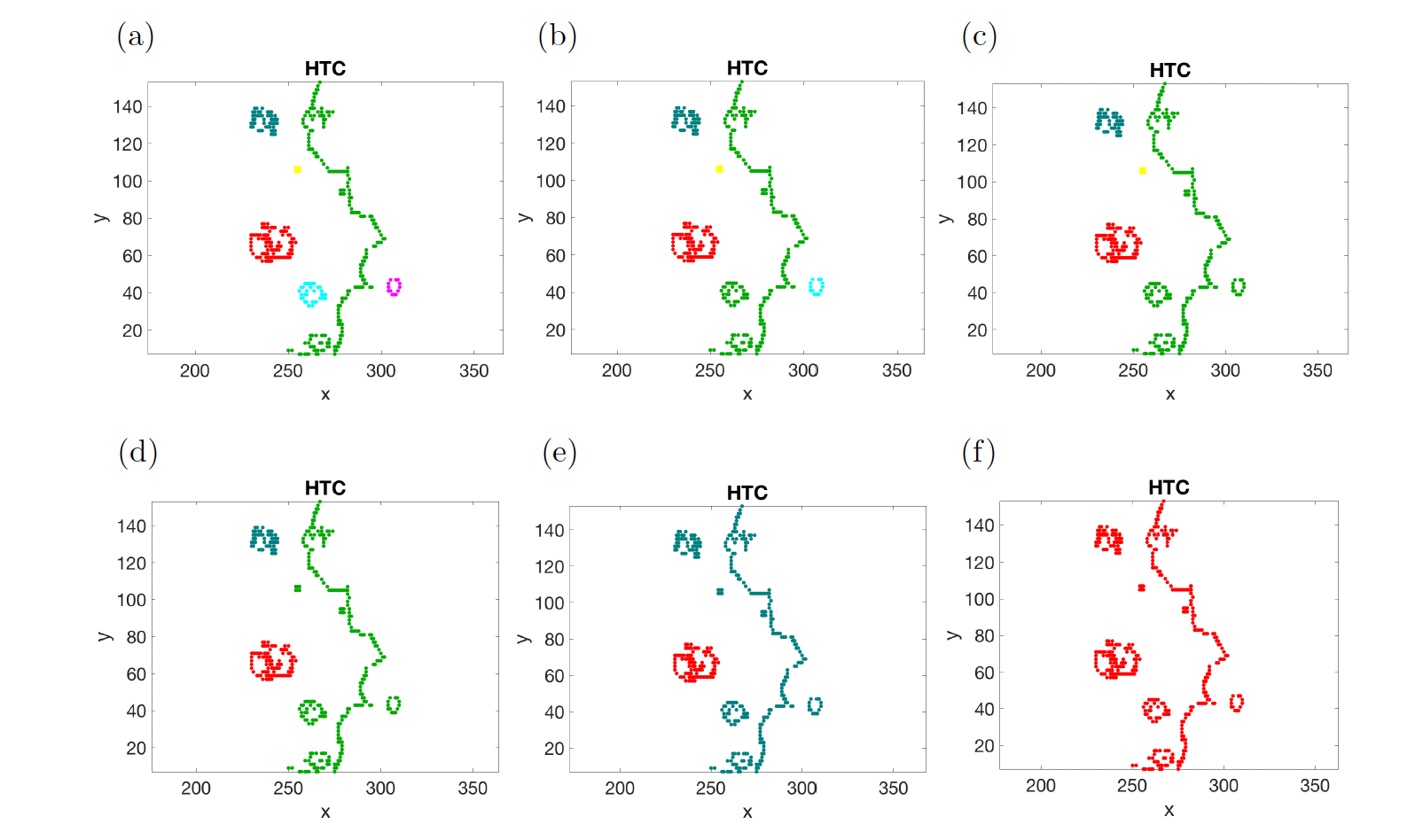}
\caption{{\bf Hierarchical topological clustering applied to a 
fragmented front separating malignant and healthy cells.}
Clusters obtained for (a) $r=5.2$, (b) $r=8.8$, (c) $r=9.6$, (d) $r=11$,
(e) $r=14.7$, (f) $r=16.9$. The last clusters to merge with the main cluster
can be considered collective outliers. They represent islands of malignant
cells that have invaded the healthy cells, while the main cluster
represents the interface between the two populations.}
\label{fig2}
\end{figure}

\begin{figure}[!hbt] \centering
\includegraphics[width=16cm]{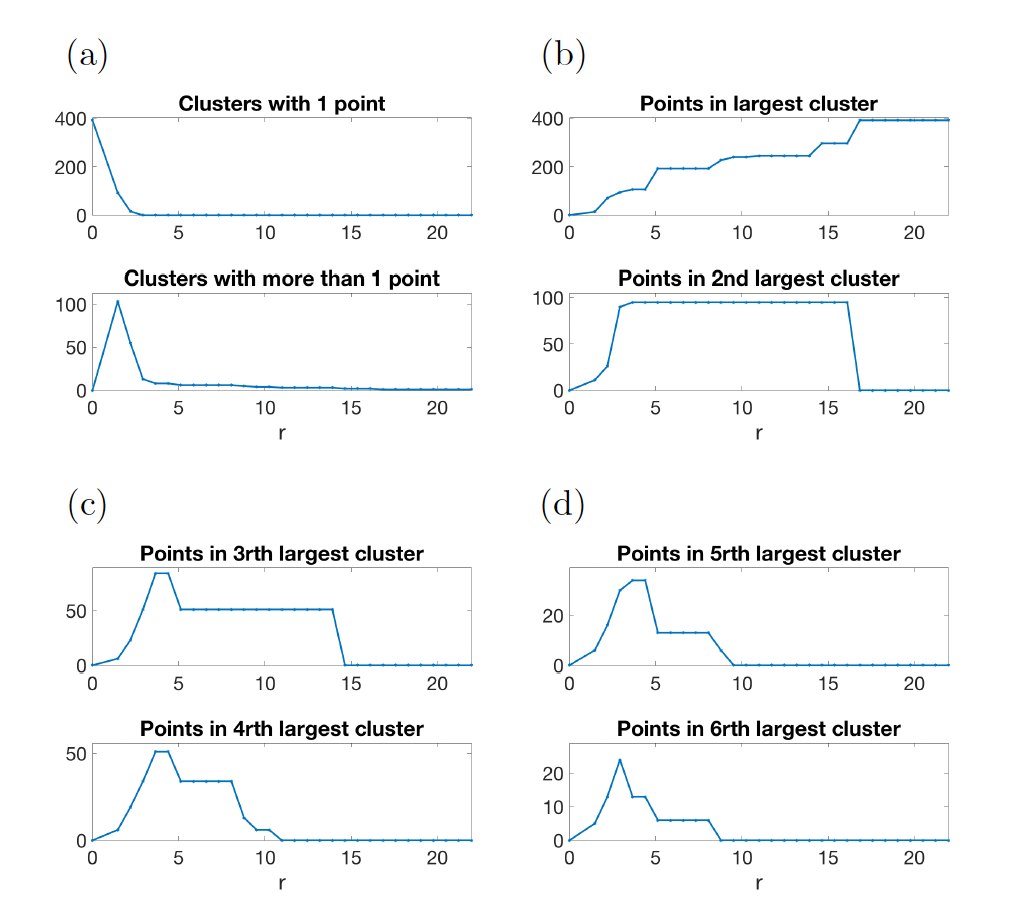}
\caption{{\bf Evolution of the topological clusters with the filtration 
parameter $r$.}
We can identify which type of clusters persist for long ranges of $r$.}
\label{fig3}
\end{figure}

\paragraph{Clustering fragmented fronts.}
%\label{sec:fronts}
Consider a two dimensional example with the Euclidean distance.
The points in Figure \ref{fig2}(f) define the interface between two types of 
cells, healthy and malignant, as extracted from an image of epithelial tissue 
\cite{plos,silberzan}. We apply the HTC strategy described in the previous 
section to the dataset formed by them.
Panels (a)-(e) illustrate stages of the evolution of the corresponding hierarchy 
of topological clusters as the filtration parameter grows. Initially, each point 
is a cluster (not depicted). 
As the filtration parameter $r$ increases, we distinguish a main cluster 
representing the interface between two different types of cells (in green)
and additional clusters representing detached islands as one 
population invades  the other, see panel (a). 
Increasing $r$ further, the closest fragments join the main cluster, until 
a single cluster is left. The last cluster to join is the one formed by 
malignant cells that have penetrated deeper  into the layer of healthy cells. 
Figure \ref{fig3} illustrates the evolution of the hierarchy of clusters as the 
filtration  parameter $r$ grows, and the persistence and size of the dominant 
clusters. In view of these graphs, we can choose filtration values of interest,
for which a number of relevant clusters are observed or relevant isolated
outliers persist. Clusters obtained in this way have an interpretation as
groups of either healthy or malignant cells and interfaces between them.
The hierarchy level at which they merge with the dominant cluster
(that initially represents the main interface between the two types of
cells) provides information on how deep the malignant cells have 
penetrated into the healthy layer.

%{/Users/anacarpio/Top-data/javaplex_matlab/data_synthetic_htda/fronts/

\begin{figure}[!hbt]
\centering
\includegraphics[width=16cm]{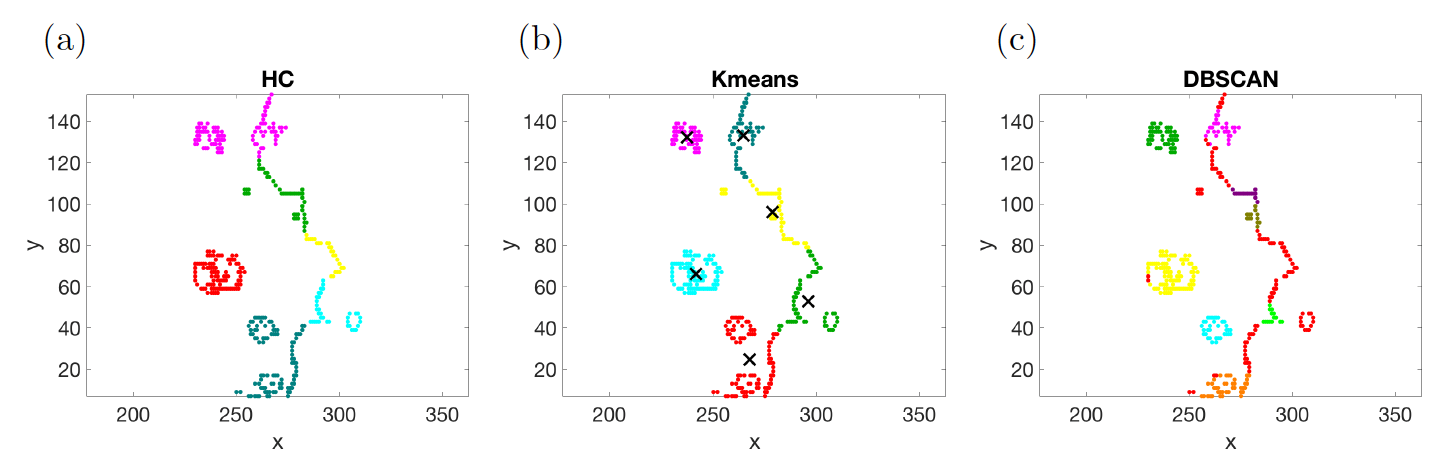}
\caption{{\bf Hierarchical clustering, K-means and DBSCAN applied
to the fragmented front.}
Color representation of (a) the six clusters generated by hierarchical
clustering with average linkage and (b) one of the six cluster arrangements
obtained by K means (different cluster arrangements are possible).  
Panel (c) displays the outcome of DBSCAN for $M_p=10$ and $\varepsilon=5$, 
red points are not assigned to any cluster since they do not have enough 
neighbors close enough.
Compare to the cluster distribution represented in Figure 1(a).
The geometrical interpretation of the clusters as representing the main 
interface and detached islands provided by HTC is lost.}
\label{fig4}
\end{figure}

While the clusters obtained by HTC have a clear meaning in this
context, this is not the case for other standard clustering methods.
Figure \ref{fig4}(a)-(b) compares to clusters obtained by K-means and
hierarchical clustering with average linkage, which fail to capture
the geometrical interpretation discussed above. Average linkage
is the choice with highest cophenetic correlation coefficient
for the data under study, compared to other standard linkage 
strategies such as single, complete and weighted linkage.
Panel (c) shows the outcome of the DBSCAN algorithm for a specific 
choice of the minimum distance $\varepsilon$ to join points and the 
minimum number of points $M_p$ to form a cluster. Red points are 
not assigned to any cluster for this choice of hyperparameters.
By trial and error, we find other choices of the hyperparameter 
parameters providing cluster distributions similar to the cluster 
arrangements automatically obtained by HTC and represented 
in Figure \ref{fig2}.  

\begin{figure}[!hbt]
\centering
\includegraphics[width=16cm]{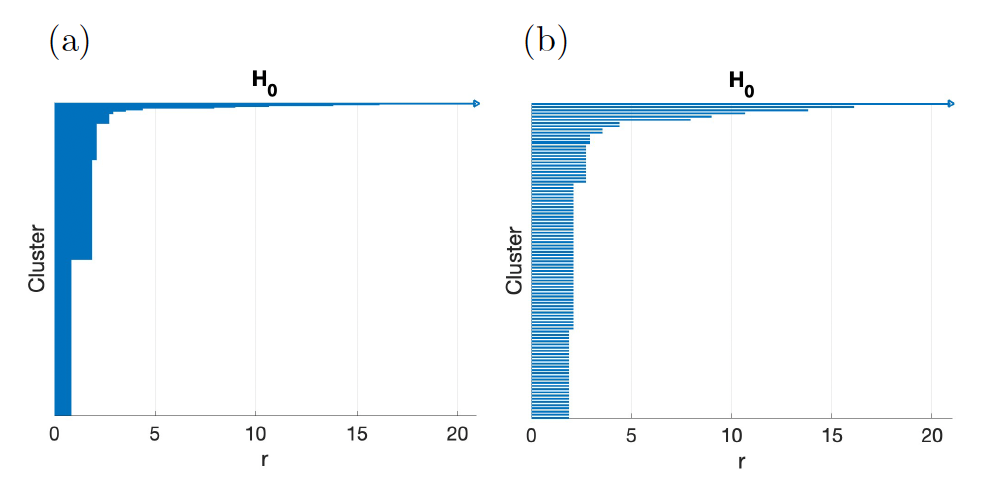}
\caption{{\bf Persistence barcode for the $H_0$ homology of the 
fragmented front dataset}.
Full barcode (a) versus amplified view of the less persistent clusters (b).
This representation illustrates the dynamics of clusters as the filtration parameter
varies in a blind way: we do not know which elements belong to each cluster.}
\label{fig5}
\end{figure}

 However, it is hard to foresee which parameter choices would 
produce DBSCAN clusters and outliers describing the interfaces
between healthy and malignant cells and detached fragments of
malignant cells invading healthy tissue.
Instead, the barcode in Figure \ref{fig5} clearly indicates the achievable
cluster numbers and the range of filtration values at which we 
expect a defined number of clusters. Our algorithm complements
information on the topological structure of the data provided by
the barcode with the knowledge of which points form each cluster
at each filtration value.

%with a geometrical meaning in this setting.

\begin{figure}[!hbt] 
\includegraphics[width=13.5cm]{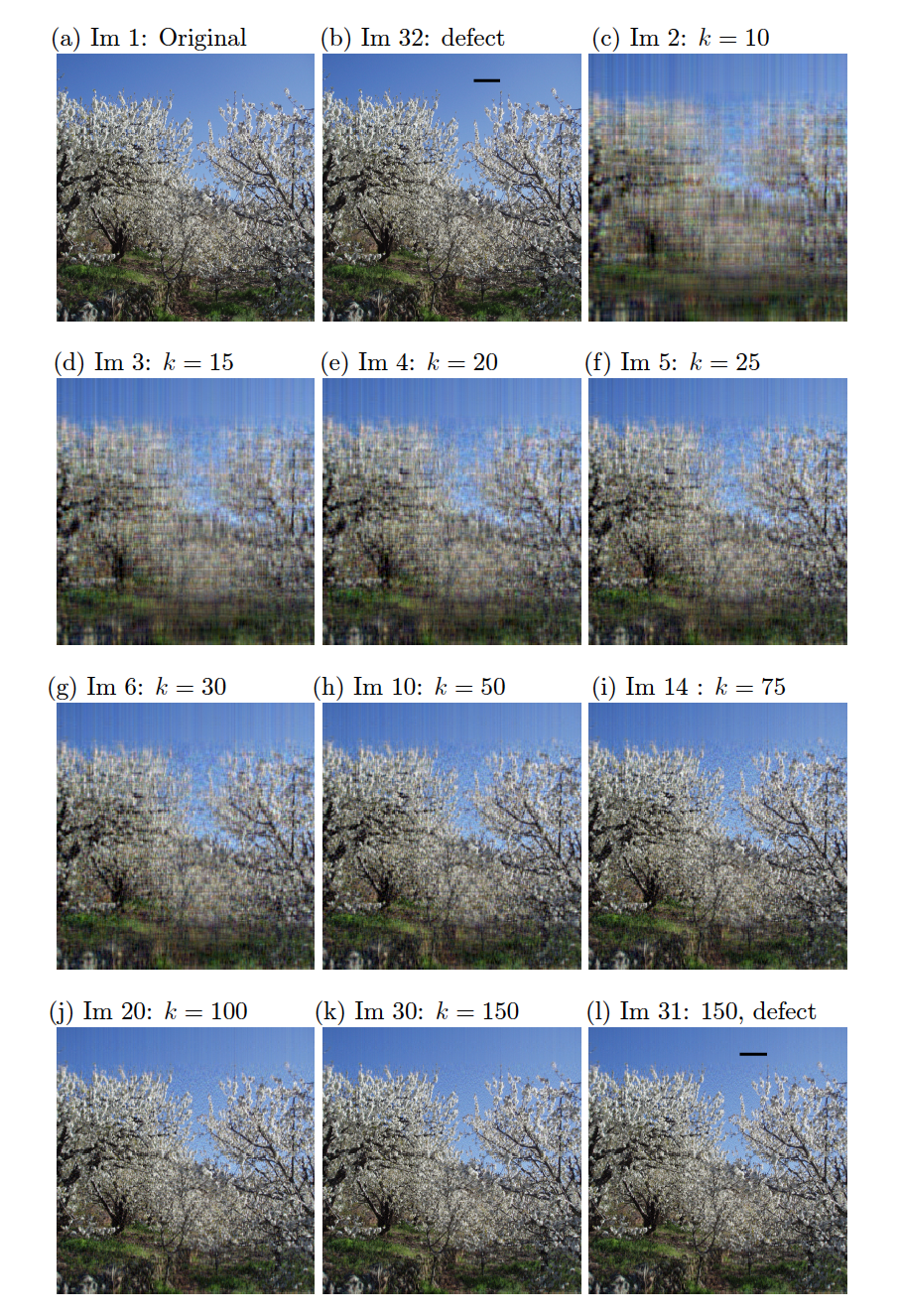}\caption{{\bf Series of images with decreasing compression and eventual defects for the image test}.
The parameter $k$ indicates the number of singular values considered in the
compressed images.}
\label{fig6}
\end{figure}

\paragraph{Quality assessment in image processing}
%\label{sec:images}
Let us test our procedure now for quality assessment in image processing.
We consider a series of images generated by slowly varying the 
compression of an image. Figure \ref{fig6}(a)-(l) presents some elements 
of the image series. Panel (a) represents the original image (image 1),
whereas panel (b) includes an additional object: a black line (image 32).
Panels (c)-(l) correspond to the original image with a compression
degree indicated by the parameter $k$, the number of singular values
kept in the compressed images \cite{compression}.
It takes the values $k=10+5 q$, $q=0,1\ldots, 28$ for images 2 to 30. 
Panel (l) includes an additional black line in Image 30 to obtain image 31. 
We compare the  resulting $32$ images using the Wasserstein distance 
$W^{1,2}$ for images \cite{wasserstein}, see Appendix for 
details. This distance is particularly suitable for image comparison because
it compares patterns, unlike Euclidean distances, which compare local
deviations.

\begin{figure}[!hbt] 
  \centering
\includegraphics[width=11.3cm]{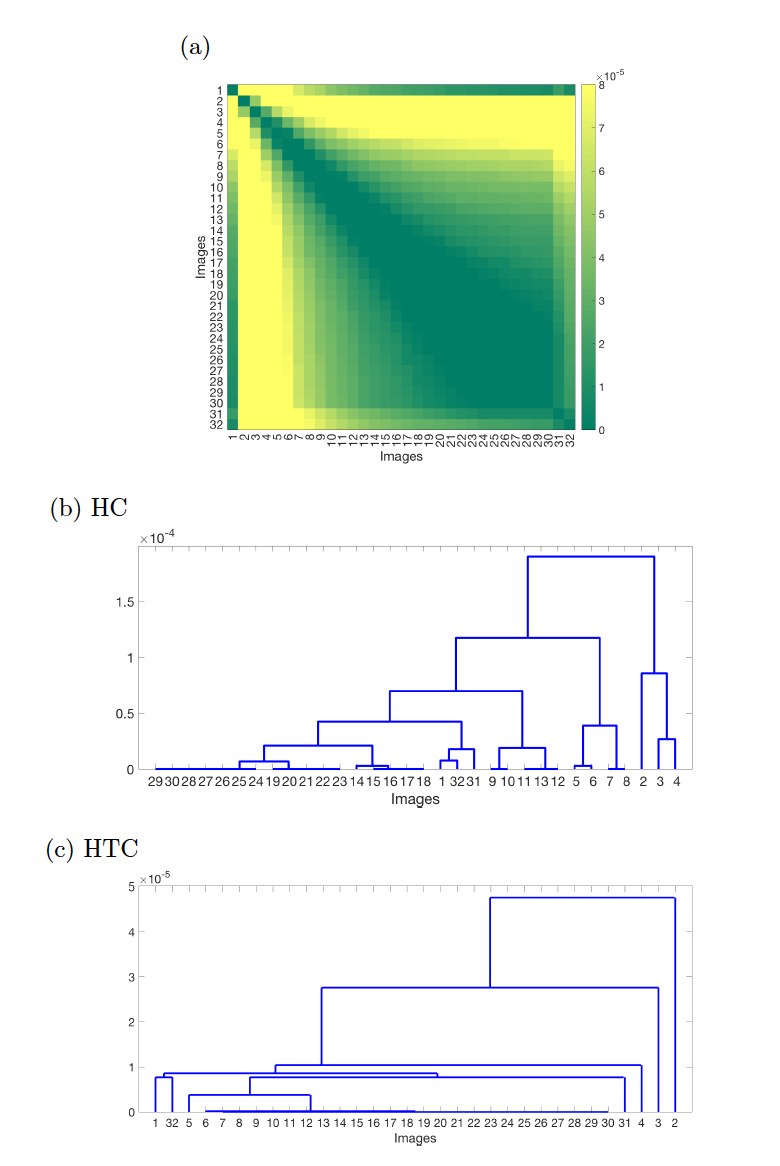}
\caption{{\bf Hierarchical clustering with the highest cophenetic correlation 
coephicient versus topological hierarchical clustering for the image test.}
Wasserstein distance matrix (a) and  dendrograms calculated by hierarchical 
clustering with complete linkage (b) and topological hierarchical clustering (c). 
While (c) distinguishes a cluster formed by uncompressed images 1 (original)
and 32 (original with line defect) from the cluster of compressed images 5-30
(without defect) and 31 (with line defect), (b) classifies images 1, 31 and
32 in the same cluster.}
\label{fig7}
\end{figure}

\begin{figure}[!hbt] \centering
\includegraphics[width=12cm]{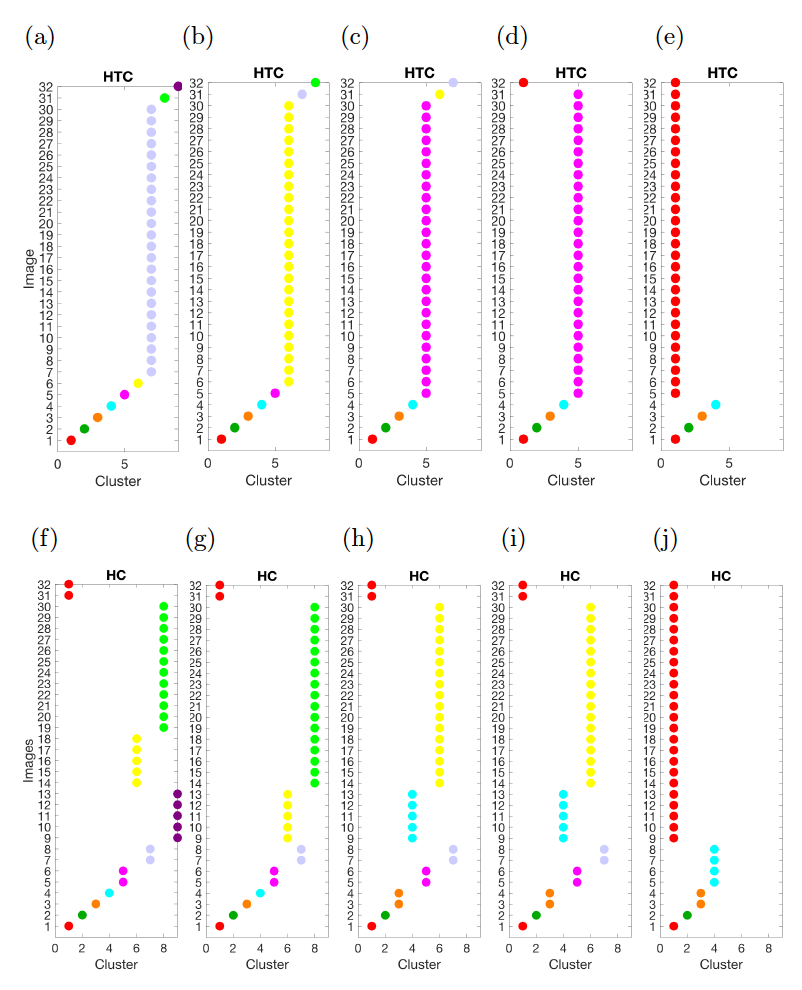}
 \caption{{\bf Hierarchical clustering with complete
linkage versus topological hierarchical clustering for the image test.}
Groupings of images obtained by topological hierarchical clustering (a)-(e)
and by standard hierarchical clustering (f)-(j) fixing a different number of
clusters. HTC forms a main cluster of similarly compressed images, but is
able to identify a secondary cluster (red) formed by the two uncompressed 
images in (d). The uncompressed (32) and compressed (31) images containing 
line  defects are classified with the other uncompressed (red) and compressed
(magenta) images in (d). Instead, HC groups these two images with the 
original image 1, see red cluster in (f)-(i), when employing the linkage 
method which displays the highest cophenetic correlation coefficient for
this dataset.}
\label{fig8}
\end{figure}

\begin{figure}[!hbt]
\centering
\includegraphics[width=12cm]{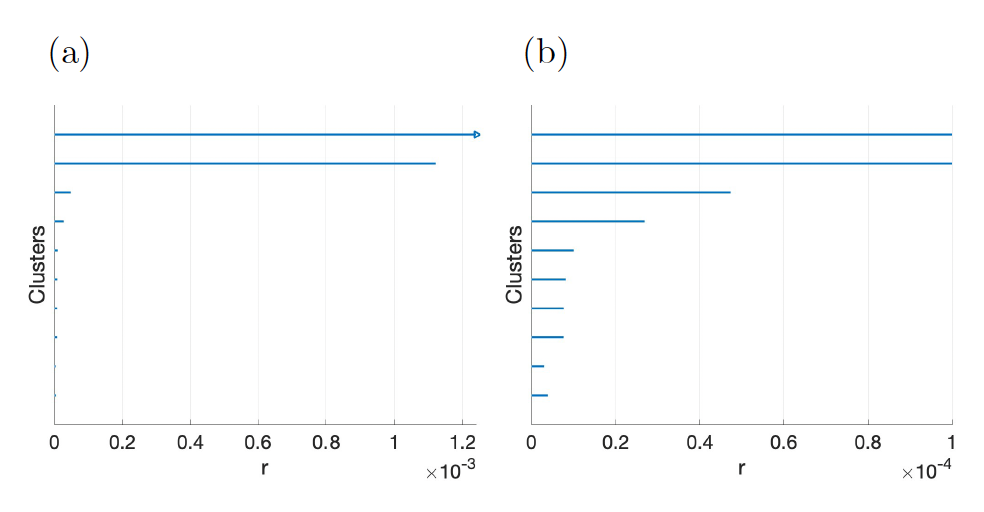}
\caption{{\bf Persistence barcode for the $H_0$ homology of the
fragmented front dataset}.
 Full barcode (a) versus amplified view of the initial clusters (b).}
\label{fig9}
\end{figure}

The heatmap in Figure \ref{fig7}(a) represents the matrix of Wasserstein 
distances. Applying the standard hierarchical clustering algorithm with 
average linkage we obtain the dendrogram represented in Figure \ref{fig7}(b).
Figure \ref{fig8}(f)-(j) visualizes how the images group in $4$, $5$, $7$, $9$ 
clusters. Topological hierarchical clustering yields different cluster and 
outlier arrangements, as illustrated by the dendrogram  in Figure 7(c) 
and panels (a)-(e) in Figure \ref{fig8}. Figure \ref{fig9} reproduces the 
dynamics of clusters as the filtration values increases. 
For adequate choices of the filtration parameter, we find the 9 groups
indicated in panel 8(a). The original picture (image 1), and the two pictures
containing a black line (images 31 and 32) are outliers, same as the
highly compressed images 2-6. The remaining images form one cluster
of recognizable images with admissible compression and no line defects.
Increasing the filtration parameter, some compressed images join
the main cluster, see panels 8(b)-(c). Next, the two images with 
black lines merge with different clusters in panel 8(d). The uncompressed 
one  forms a group with the original image, while the compressed one joins
the main cluster of compressed images. 

% si meto las letras en las figuras, ocupará menos

Figure \ref{fig7}(c) shows that these
two clusters, the one formed by the uncompressed images 1 and 32,
and the moderately compressed images 5-31 persist for a range
of filtration values. It also shows that the images with line defects are
distinguished from the other images with similar compression since
they merge with them for higher values of the filtration value.
These two clusters merge in panel 8(e) leaving only the more blurry 
images as persistent outliers. 
This hierarchy allows us to single out in a automatic way images
corrupted by the addition of extra elements (the line) and images with
poor resolution (too compressed). 
The fact that HTC classifies initially the compressed image 31 with the 
compressed images without a line, and uncompressed image 32 
with uncompressed image 1, reflects that the distortion of the overall 
image features caused by the selected levels of compression has an 
stronger weight on the distance between the images forming the dataset 
that the presence of the line. Many details of flowers, branches, soil
and sky are blurred over wide areas.
Instead, the clusters obtained by hierarchical clustering with complete
linkage represented in Figures \ref{fig7}(b) and \ref{fig8}(f)-(j) focus more 
on the compression  level. Complete linkage attains the highest cophenetic 
correlation coefficient for this dataset.
The compressed and uncompressed images 31 and 32 containing 
line defects are already classified with the original image 1 in panel 8(f).
Topological hierarchical clustering may become a useful tool  for 
image quality control, when monitoring the appearance of sudden local
changes in the image.

\paragraph{Outliers in economy datasets}
%\label{sec:economy}
Outliers are particularly meaningful in social sciences and economy.
We can use these techniques to  detect  the best candidates for a job,
for instance.
We consider here data from the trade statistics for international business 
development \cite{edata}. This open access database contains
trade statistics for international business development. It includes
monthly, quarterly and yearly trade data from all countries:
import and export values, volumes, growth rates, market shares...
We consider in particular the export and import values for
all products (about 100 categories) in trade between a fixed country, 
Spain, and all countries in the European zone. Consider data
from $43$ countries during the year $2019$, for example.

\begin{figure}[!hbt] \centering
\includegraphics[width=11.2cm]{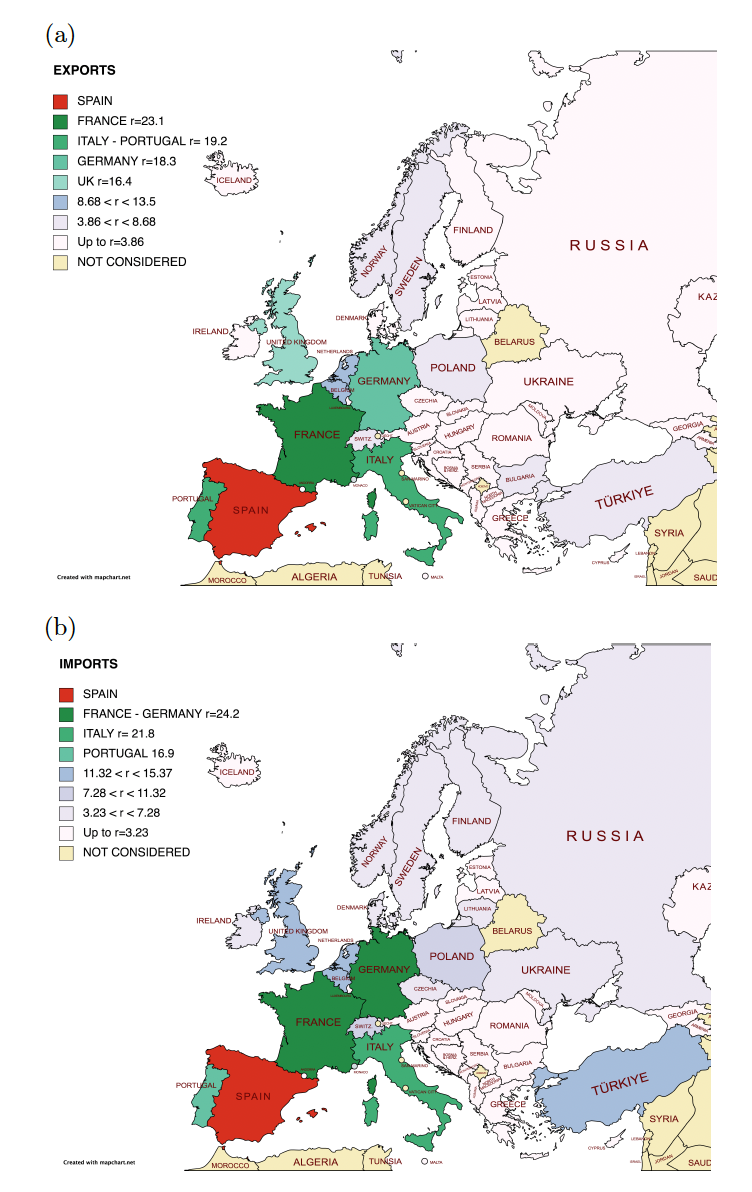}
  \caption{{\bf Relevant and irrelevant partners in 2019 Spanish trade.}
Cluster of low interaction and dominant partners (outliers) in Spanish (a) 
exports and (b) imports at European level,  representing the main providers 
and receivers for Spanish economy.
Size and proximity seem relevant factors. Visualized by
https://www.mapchart.net/.
Colors represent ranges of the filtration parameter $r$ for which the colored 
countries merge with the main cluster. $r$ increases as the level of exports 
(resp. imports) becomes more relevant.}
\label{fig10}
\end{figure}

First, we normalize the values $g_{n,m}$ for each product $m$, 
$m=1,\ldots,M$ subtracting the  mean $\mu_m$ over $N$ countries 
and dividing by the standard deviation $\sigma_m$:
$\tilde g_{n,m} = (g_{n,m} - \mu_m)/(3 \sigma_m)$,
$n=1,\ldots,N$,  $m=1,\ldots, M$.
Next, we calculate the  Euclidean distance between the countries
$c_n$ and $ c_k$
using either the export or the import normalized values $\tilde g_{n,m}$:
$d(c_n,c_k)= (\sum_{m=1}^M | \tilde g_{n,m} - \tilde g_{k,m} |^2)^{1/2}$ 
$n, k=1,\ldots,N$. Applying the HTC algorithm, we end up with the
outliers indicated in Figure \ref{fig10}. In both cases most 
countries form a big cluster while a few dominant outliers persist and merge 
with it sequentially as the filtration parameter $r$ that governs the hierarchy
increases. 

Regarding exports from Spain to other European countries in 2019,
a first cluster of countries which receive almost no exports arises, 
formed by 18 countries for low $r\sim 10^{-1}$:
Albania, Armenia, Bosnia, Georgia, Iceland, Kazakhastan, Macedonia,
Moldova, Montenegro, Cyprus, Estonia, Latvia, Lithuania, Malta, 
Serbia, Slovenia, Luxembourg, Ukraine. At $r=3.8$, this cluster of
low level importers is formed by $30$ countries colored in light blue in
Figure \ref{fig10}(a). The main importers of Spanish products are persistent
outliers that merge with the main cluster at high values of $r$.
France arises as the main importer, with Italy and Portugal basically 
tied up in second position and Germany third.  

When we look at Spanish imports, the initial cluster from which there are
almost no imports is formed by $14$ countries for low $r \sim 10^{-1}$: 
Albania, Andorra, Bosnia, Montenegro, Macedonia, Croatia, Cyprus, 
Georgia, Estonia, Latvia, Malta, Moldova, Serbia, Slovenia. At 
$r=3.2$, this cluster of low level exporters contains $24$ 
countries represented in light blue in Figure \ref{fig10}(b). Germany and 
France arise as dominant exporters to Spain, followed by Italy. 
Both Italy and Portugal in the previous case, and Germany and France 
in this case, form clusters of two items before merging with the main one.

When using K-means, the choice of the number of clusters may be 
unnatural, and force countries to be together in an arbitrary way. The 
hierarchical clustering procedure with the highest cophenetic correlation 
coefficient corresponds to average linkage, which defines intercluster
distance as an average of the distance between their points. Differences
arise in the order the countries join the main clusters. For instance,
while HTC clearly distinguishes Germany and France as main import
partners keeping them isolated from the rest until the top filtration
values, Italy is last to merge the main cluster using HC with average 
linkage. The interpretation of this fact using average distances is obscure,
unlike the interpretation of separate connected components.

\paragraph{Analysis of gene data}
%\label{sec:gene}
Gene analyses provide challenging datasets to test clustering procedures.
We consider here data from  the cBio Cancer Genomics open access
Portal  \cite{cbioportal}, more precisely, mRNA gene expression data 
from the {\it Breast invasive carcinoma} TGCA Pancancer Atlas dataset  
\cite{pancancer,tgca_breast}.
This dataset contains measurements of the mRNA gene expression
of  $20531$ genes  for $1082$ cancer samples and also for a set for 
$114$ healthy samples. 

\begin{figure}[!hbt] \centering
\includegraphics[width=16cm]{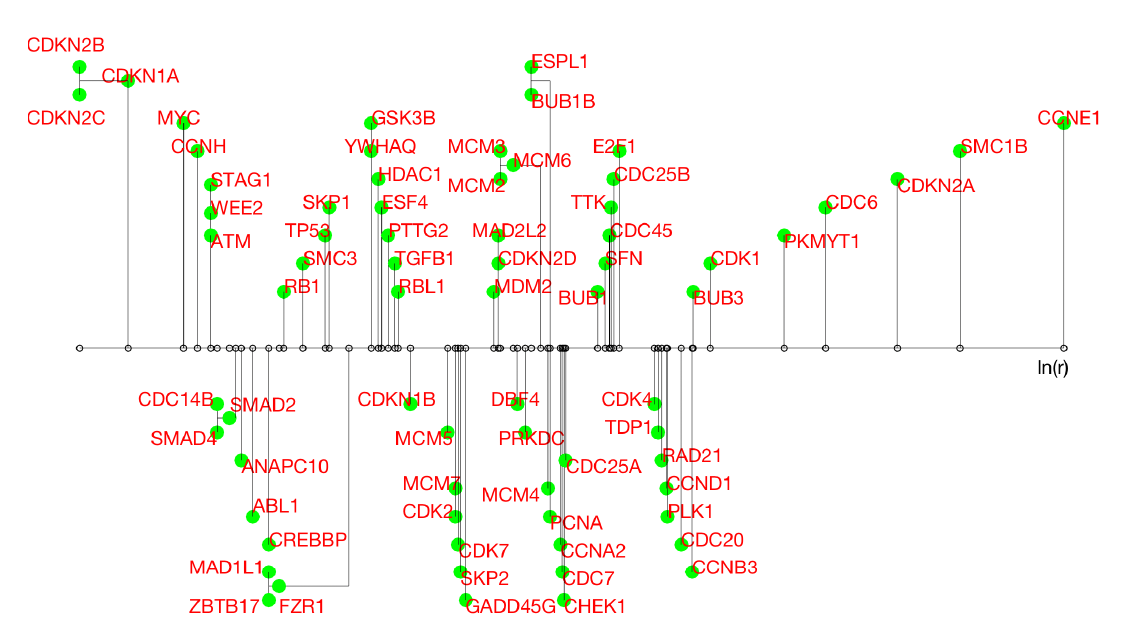}
\caption{{\bf HTC study of mRNA gene expression using TGCA breast cancer data 
for genes involved  in the cell cycle.} 
The scheme lists the genes in the cell cycle as they join the main HTC cluster while 
increasing the filtration parameter $r$. The last genes to merge are persistent outliers 
and are believed to play relevant roles in cancer processes. Notice the logarithmic
scale: most genes collapse to one cluster while a few dominant outliers persist.}
\label{fig11}
\end{figure}

\begin{figure}[!hbt]
\centering
\includegraphics[width=8cm]{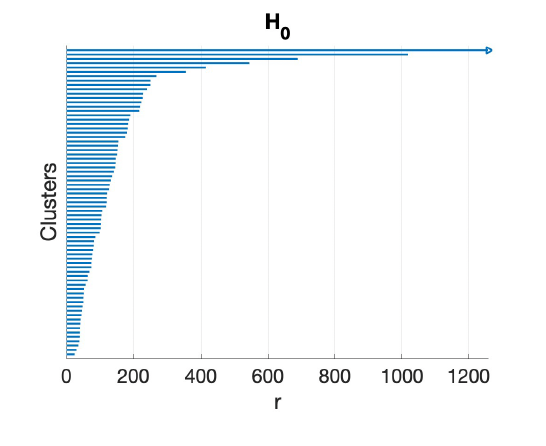}
\caption{{\bf Persistence barcode for $H_0$ homology of the mRNA gene expression 
data for genes involved  in the cell cycle}.}
\label{fig12}
\end{figure}

We are interested in analyzing gene expression in the cancer samples. 
Since the order of magnitude of the measured values can largely vary
for each gene, it is convenient to normalize the data for cancer samples 
using some reference values. A standard procedure employs the data 
from  all the available healthy samples to define means and standard 
deviations for the  expression of each gene in healthy tissue, which are 
then exploited to normalize data from cancer tissue.
For each gene $m$, we calculate the mean  $\mu_m$  and  the standard 
deviation $\sigma_m$ of the mRNA gene expression values over the 
$114$ available healthy samples.
Then, we normalize the expression values $g_{n,m}$ of each gene $m$
for each cancer sample $n$ subtracting the  mean $\mu_m$  and 
dividing by the standard deviation $\sigma_m$:
$\tilde g_{n,m} = (g_{n,m} - \mu_m)/(3 \sigma_m)$,
$n=1,\ldots,N$,  $m=1,\ldots, M$.
Next, we calculate the  Euclidean and the Fermat distance \cite{fermat}
between the  normalized gene expressions $\tilde g_{n,m}$, see 
Appendix for details. We extract values for $M=71$ genes  
%\footnote{ \small
 %ABL1, ANAPC10, ATM, BUB1, BUB1B, BUB3, CCNA2, CCNB3,
% CCND1, CCNE1, CCNH, CDC14B, CDC20, CDC25A, CDC25B,
% CDC45, CDC6, CDC7, CDK1, CDK2, CDK4, CDK7, CDKN1A,
% CDKN1B, CDKN2A, CDKN2B, CDKN2C, CDKN2D, CHEK1,
% CREBBP, DBF4, E2F1, E2F4, ESPL1, FZR1, GADD45G, GSK3B,
 %HDAC1, MAD1L1, MAD2L2, MCM2, MCM3, MCM4, MCM5, MCM6,
% MCM7, MDM2, MYC, PCNA, PKMYT1, PLK1, PRKDC, PTTG2,
% RAD21, RB1, RBL1, SFN, SKP1, SKP2, SMAD2, SMAD4, SMC1B,
% SMC3, STAG1, TFDP1, TGFB1, TP53, TTK, WEE2, YWHAQ, ZBTB17.}.
 %Absent from the TGCA study (6):
 %ORC1, ORC2, ORC3, ORC4, ORC5, ORC6.}.
involved in the cell cycle gene/protein interaction network \cite{cell_cycle}.
Applying the HTC algorithm to the cancer samples we find  sets 
of distinguished outliers, such as CCNE1, SMC1B, CDKN2A, 
CDC6,  PKMYT1, CDK1. All of them have been proven to be relevant
for breast cancer prognosis or singled out as anti-cancer therapeutic 
targets  \cite{cancer1,cancer2,cancer3,cancer4,cancer5,cancer6}.
Figure \ref{fig11} lists the genes we have considered in the order they
merge with a dominant cluster as the filtration parameter $r$ grows
in a logarithmic scale. Figure \ref{fig12} displays the standard $H_0$
homology plot. Notice that the genes merge the main cluster almost
sequentially. The last genes to merge are distinguished outliers.

\begin{figure}[!hbt]
\centering
\includegraphics[width=16cm]{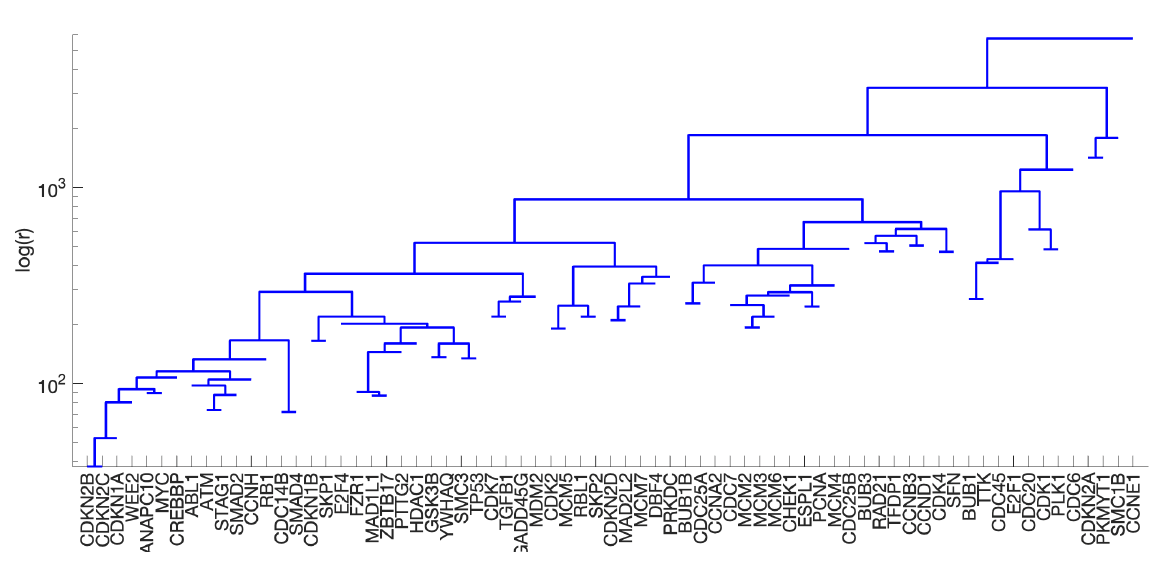}
\caption{{\bf Hierarchical clustering with weighted linkage applied to
mRNA gene expression  data for genes involved  in the cell cycle} 
Compared to the sequential merging process observed in HTC intermediate 
groups are formed.}
\label{fig13}
\end{figure}

The hierarchical clustering procedure with the highest cophenetic 
correlation coefficient corresponds to weighted linkage, which defines 
intercluster distance in a recursive way, by averaging the distances
between clusters that have merged to creare it before, see Appendix. 
This method provides a dendrogram depicted in Figure
\ref{fig13} in logarithmic scale. Compared to Figure \ref{fig11},  it stresses
intermediate subclusters of unclear meaning in this context, which can be 
seen in HTC as groups of genes collapsing at similar filtration levels, but 
dilutes the dominant meaningful outliers.

%\section*{Discussion}

\section{Conclusions}
%\label{sec:conclusions}

We propose a hierarchical clustering algorithm based on elementary
persistence homology calculations. Once a distance function is selected, 
a threshold value for the distance between points controls cluster
formation. All elements in a cluster can be joined by a path of elements
at a distance smaller than the threshold from one to the next. The resulting
clusters can take any shape. As the threshold value increases, existing
clusters merge whenever there are two elements in them which become
close enough. This shape based procedure selects persistent outliers as 
elements which are far from all the rest. For them, it takes large threshold 
values to merge with an existing cluster. Our approach is different
from other topological studies that calculate and compare persistent diagrams 
through Bottleneck or other distances and then employ standard hierarchical
clustering algorithms.

Our procedure enjoys geometrical interpretations which are absent
in other standard clustering procedures. We have evidenced this
fact comparing with the groupings provided by different clustering
strategies when studying a fragmented interface separating two
types of cells. While the hierarchical topological clustering method
we propose distinguishes automatically the main interface and detached 
islands, k-means, standard hierarchical clustering, and even blind 
DBSCAN fail to do so. An additional study of a collection of images, 
shows that  HTC is able to distinguish compressed and uncompressed 
images, with defects and without defects, in situations where HC fails
to do so. Our procedure is able to detect meaningful collective outliers 
(detached islands in the fragmented interface test) as well as single 
outliers (images that have been excessively compressed). We have 
illustrated its usefulness for the study of trade data (creating a 
hierarchy of importers and exporters that visualizes the key partners) 
and cancer gene data analysis (identifying persistent outliers whose role 
may deserve specific studies).  \\

% Cancer - General
% Economics
% Computer and information sciences - Artificial intelligence, machine learning and data science

{\bf Acknowledgements.}
Research supported by Spanish MICINN grants PID2020-112796RB-C21 and
PID2024-155528OB-C21.

{\bf Author contributions.}
 AC designed research, algorithm and code. AC and GD performed data studies.
 AC wrote the paper. All authors read and approved the final manuscript.  
 
{\bf Availability of data and material.}
Open access data for economy studies are available from ITC Market Analysis Trade Map https://www.trademap.org.
Open access data for breast cancer gene studies are available from the cBioPortal from Cancer Genomics at
http://www.cbioportal.org/study/summary?id=brca\_tcga\_pan\_can\_atlas\_2018.
The data for images and fronts can be extracted from the Manuscript images.
Codes are implemented in Matlab and can be available upon reasonable request.

{\bf Conflict of interest.}
 The authors declare that they have no competing interests.

\appendix

\section{Clustering algorithms} 

We briefly summarize the basis of the clustering algorithms used for comparison.

{\it K-means} clusters data in groups in order to minimize the total 
intra-cluster variation, which measures the cluster compactness \cite{kmeans}. 
Given a data cloud $\mathbf x_i=(x_{i,1}, \ldots, x_{i,M})$ in a $M$ dimensional 
space, the intra-cluster total variation is given by
$ \sum_{j=1}^{J} W(C_j) = \sum_{j=1}^{J} \sum_{\mathbf x_i \in C_j} d(\mathbf x_i,
\boldsymbol \mu_j),$
where $C_j$ is a cluster of such points and $\boldsymbol \mu_j$ is
the cluster centroid. Each term $W(C_j)$ represents the variation 
within a cluster.  Here, the distance $d$ stands for the Euclidean distance 
$d(\mathbf x_i,\boldsymbol \mu_j)^2 =  \sum_{m=1}^M (x_{i,m} - \mu_{j,m})^2.$
The K-means algorithm proceeds in the following steps. We fix the number of
clusters $k$ to be formed and initialize the centroids $\boldsymbol \mu_j$ 
by randomly generating $k$  points. Next, each datum $\mathbf x_i$ is assigned 
to the centroid minimizing the Euclidean distance. Within each cluster, we set
the average of the cluster points as the new centroid. These steps are repeated 
until the clusters do not change.
K-means  needs one hyperparameter to proceed:  the number of clusters. 
There  are specific criteria such as the Elbow or Silhouette methods to propose 
a tentative cluster number.

{\it Hierarchical clustering} produces a multilevel hierarchy, in which 
clusters  at one level coalesce at the next level \cite{clusters}.  The agglomerative 
algorithm starts from as many clusters as data points.  Nearby clusters merge to 
create larger ones until all the data points form a single cluster. This procedure is 
schematized in a dendrogram, a graph visualizing how the clusters join
until  they form a tree that comprises all, see Figure 5(b). 
To evaluate the proximity of clusters and merge them, the algorithm employs
`linkage functions': `single', `average', `complete', `weighted', `centroid', `median',  
`ward', though `centroid', `median', and `ward' tend to be used with Euclidean
distances.
Both the linkage functions and the distance are hyperparameters to be 
selected. Here, we have chosen different distances depending on the
test case and implemented the linkage providing the biggest cophenetic 
correlation coefficient for these datasets, that is, the biggest correlation 
between the original distance and the cophenetic distance (the height at 
which clusters coalesce). Large correlation  indicates that the tree is 
representative of our dataset. The remaining hyperparameter, 
that is, the height, determines the number of clusters. Cutting the tree at 
different  heights, we select specific numbers of clusters.
In the examples discussed, the linkage methods displaying highest cophenetic
correlation coefficients are `complete' ,  `average' or `weighted', depending on
the dataset considered.
Given two clusters with elements $x_{j_1},\ldots,x_{j_J}$ and $x_{i_1},\ldots,x_{i_I}$,
complete linkage defines the distance between clusters as the maximum distance
between elements, that is,
${\rm max}\{d(x_{j_k},x_{i_\ell}), \, k=1,\ldots,J, \, \ell=1,\ldots, I\}$.
Average linkage defines the distance between clusters as
${1\over IJ}\sum_{k=1}^J \sum_{\ell=1}^I d(x_{j_k},x_{i_\ell})$.
Weighted linkage defines the distance between clusters in a recursive way.
Assume we create cluster $C$ by merging smallest clusters $C'$ and $C''$. Then
the distance between another cluster $Q$ and $C$ is the average of the distances
between $Q$ and $C'$ and between $Q$ and $C''$: 
$d(C,Q)= {(d(C',Q)+d(C'',Q))/2}.$

{\it Density-Based Spatial Clustering of Applications with Noise} (DBSCAN) 
algorithms  \cite{dbscan}  define clusters in high density regions, leaving observations in low density regions outside, which eventually become anomalies. 
The process starts with an arbitrary data point that has not yet been classified. We
find the points at a distance smaller than $\varepsilon$ (the $\varepsilon$-neighbourhood).
When it contains more than a minimum number of points $M_p$, we create a new 
cluster with them.
Otherwise, we consider that point as noise. However, this point might become part
later of the $\varepsilon$-neighbourhood of a different point containing enough points, and, thus, belong to that cluster. If not, it remains an outlier. Thus, this algorithm can identify  non convex clusters and outliers. However, finding good values  for the two hyperparameters $\varepsilon$ and $M_p$ is a nontrivial task, strongly dependent on the dataset's structure.  In principle, the distance has to be selected too, but we have fixed the Euclidean distance.

\section{Distances}
 
Here we describe some distances used to treat images and data.

{\it Wasserstein Distances.}
Fast algorithms to calculate Wasserstein-1 distances between distributions 
defined on a grid are proposed in \cite{wasserstein}. Their calculation is posed  
as optimal transport  problems. Optimal transport plays crucial roles in many 
areas, including  image processing and machine learning.
Given two 2D probability distributions, or two images,  $\rho^0$ 
and $\rho^1$, they seek a transport plan $f(\mathbf x)$ from one to the other  
with minimal cost
\begin{eqnarray*}
    \mbox{min}_f \displaystyle{\int_{\mathbf x}{||f(\mathbf x)||_p \ d\mathbf x}} \label{wd}
\end{eqnarray*} 
such that ${\rm divergence}_h(f) = \rho^0-\rho^1$ under a zero-flux boundary 
condition. Here, $p$ can be 1,2 or infinity, so that $||f(\mathbf x)||_p$ are the 1,2 or 
infinity norms of $f(\mathbf x)$, respectively,  h is the grid step size and
${\rm divergence}_h$ is a divergence operator defined on the grid,
see  \cite{wasserstein} for details.

{\it Fermat distances.}
Considering a set S of columns (resp. rows) $\mathbf m^1$, $\mathbf m^2$, \ldots, 
$\mathbf m^L$, the {\it Fermat} $\alpha$-distance between any two of them  
relative to that set is  \cite{fermat}
\begin{eqnarray*}
d_{S,\alpha}(\mathbf m^1,\mathbf m^2)=\mbox{min}\left\{\sum_{\ell=1}^{k-1}
\| \mathbf y^{\ell+1}\!-\! \mathbf y^\ell \|_2^\alpha \bigg\vert (\mathbf y^1,\ldots,
\mathbf y^k) \mbox{ path from $\mathbf m^1$ to $\mathbf m^2$ in S}\right\},  
\label{eq2}
\end{eqnarray*}
for any $\alpha >1$. When $\alpha=1$, we recover the Euclidean distance.
The Fermat distance compares items in a set weighting information from all the 
other items in the same set, which is interesting when we want to compare
gene profiles weighting information from cohorts of patients.

\section{Image Compression}
We detail here the compression procedure used to generate the images we compare.
Given a $m \times n$ matrix $\mathbf M$, its singular value descomposition takes the form $\mathbf  M = \mathbf  U \boldsymbol \Sigma \mathbf V^t$, where 
$\boldsymbol \Sigma$ is a $m \times n$ rectangular diagonal matrix with non-negative  real  numbers on the diagonal,  $\mathbf  U$ is a $m \times m$ unitary matrix and  $\mathbf V$ is a $n \times n$ unitary matrix. The diagonal elements $\sigma_{ii}$ of $\boldsymbol \Sigma$ are the singular values of the matrix $\mathbf M$, while the columns of $\mathbf U$ and the columns of $\mathbf V$ are the left-singular vectors  and right-singular vectors of $\mathbf M$, respectively. Compression strategies order the singular values in decreasing order and keep only $k$ of them in the factorization, that is, we set $\mathbf  M_k = \mathbf  U_k \boldsymbol \Sigma_k \mathbf V_k^t$,  where $\boldsymbol \Sigma_k$ contains only the $k$ largest singular values, while $\mathbf  U_k$ and $\mathbf  V_k$ keep only theleft-singular vectors and  right-singular vectors  associated to them
\cite{compression}.

%For more questions regarding reference style, please refer to the \href{http://www.ncbi.nlm.nih.gov/books/%NBK7256/}{Citing Medicine}.

\end{document}